\documentclass[conference]{IEEEtran}

\usepackage{url}
\input epsf
\usepackage{graphicx}
\usepackage{amsmath}

\usepackage{threeparttable}
\usepackage{amssymb}

\usepackage{xcolor}

\begin{document}

\title{\LARGE Predicting Winners of the Reality TV Dating Show \textit{The Bachelor}\\Using Machine Learning Algorithms}




%
\author{\authorblockN{Abigail J. Lee\authorrefmark{1}\authorrefmark{2},
Grace E. Chesmore\authorrefmark{1}\authorrefmark{3}, Kyle A. Rocha\authorrefmark{1}\authorrefmark{4}, Amanda Farah\authorrefmark{1}\authorrefmark{3}, Maryum Sayeed\authorrefmark{1}\authorrefmark{5}, Justin 
Myles\authorrefmark{1}\authorrefmark{6}
}
\authorblockA{\authorrefmark{1} Department of Reality TV Engineering, University of Chicago}
\authorblockA{\authorrefmark{2} Department of Astronomy \& Astrophysics, University of Chicago, 5640 South Ellis Avenue, Chicago, IL 60637\\}
\authorblockA{\authorrefmark{3} Department of Physics, University of Chicago, 5720 South Ellis Avenue, Chicago, IL 60637, USA\\}
\authorblockA{\authorrefmark{4} Department of Physics and Astronomy, Northwestern University, 2145 Sheridan Road, Evanston, IL 60208, USA}
\authorblockA{\authorrefmark{5} Department of Astronomy, Columbia University, 550 West 120th Street, New York, NY 10027, USA}
\authorblockA{\authorrefmark{6} Department of Physics, Stanford University, 382 Via Pueblo Mall, Stanford, CA 94305, USA\\}
Email: abbyl@uchicago.edu\\
Email: chesmore@uchicago.edu}


\maketitle 
\begin{abstract}
\textit{The Bachelor} is a reality TV dating show in which a single bachelor selects his wife from a pool of approximately 30 female contestants over eight weeks of filming (American Broadcasting Company 2002).
We collected the following data on all 422 contestants that participated in seasons 11 through 25: their Age, Hometown, Career, Race, Week they got their first 1-on-1 date, whether they got the first impression rose, and what ``place" they ended up getting. 
We then trained three machine learning models to predict the ideal characteristics of a successful contestant on \textit{The Bachelor}.
The three algorithms that we tested were: random forest classification, neural networks, and linear regression. We found consistency across all three models, although the neural network performed the best overall.
Our models found that a woman has the highest probability of progressing far on \textit{The Bachelor} if she is: 26 years old, white, from the Northwest, works as an dancer, received a 1-on-1 in week 6, and did not receive the First Impression Rose.  Our methodology is broadly applicable to all romantic reality television, and our results will inform future \textit{The Bachelor} production and contestant strategies. While our models were relatively successful, we still encountered high misclassification rates.  This may be because: (1) Our training dataset had fewer than 400 points or (2) Our models were too simple to parameterize the complex romantic connections contestants forge over the course of a season. 
\end{abstract}
\IEEEoverridecommandlockouts

%
\IEEEpeerreviewmaketitle

\section{Introduction}
\textit{The Bachelor} first aired in 2002, and in its 26-season run, has been highly influential, inspiring numerous spin-offs (e.g., \textit{Bachelor in Paradise, The Bachelorette, Bachelor Winter Games, Bachelor Pad}) and knock-offs (e.g., \textit{Love Island, Too Hot to Handle, Temptation Island, Love is Blind, Fboy Island}). \textit{The Bachelor} has been largely successful,\footnote{Relative to other reality TV dating shows} with six couples still together today. 

Here, we give a brief summary of the show's format.
Each season, $\sim$30 female contestants fight to compete for the male lead's heart over an eight-week period, with two to five contestants being eliminated each week. The show culminates in a marriage proposal to the last remaining woman. 
Perhaps the most impactful day is Night One, when contestants step out of a limousine to meet the lead during their ``limo entrance.'' 
At the end of the night, the lead gives the ``First Impression Rose" to whomever he had the most initial chemistry with (i.e. whomever he found the most attractive).  For the remainder of the show, the participants travel to romantic locations for their dates every week. One to two women each week are also picked for a ``one-on-one,'' in which the woman is granted a day-long date with the lead.  It is often postulated that receiving a one-on-one early in the season is a sign that one will make it far into the show; however, this has not yet been tested empirically. 

Race and age may also implicitly influence how far a contestant makes it. Out of all 26 seasons, only one woman of color (Filipino-American Catherine Lowe, season 17) has ever won the \textit{The Bachelor}. In addition, ``older''\footnote{Within the context of reality TV and the entertainment industry in general, old is generally defined to be $\gtrsim 30$ yrs.} women, often referred to as ``cougars'' on the show, rarely make it far (The only 30+ woman to win the show is 33-year-old Mary Delgado, season 6).

Love is not the only recurring theme of this masterful television show. Drama \textit{always} accompanies the lead's journey to find true love. Common dramatic tropes include: whether a contestant is ``there for the right reasons," a contestant secretly having a boyfriend back home, or the lead falling in love with multiple women (see \textit{The Bachelor} seasons 20 \& 26). Often, the women fight amongst themselves throughout the season, providing another nuanced dimension of entertainment.

Predicting winners for \textit{The Bachelor} (and similar reality TV shows) is of great interest not only to contestants and die-hard fans of the series, but also for monetary gain through betting \cite{BUNKER201927}. Many viewers participate in `Bachelor Fantasy Leagues,'\footnote{See \url{https://www.bachbracket.com/} for an example.} where machine learning would be greatly advantageous in strategical planning. 
With 26 seasons and 30 contestants per season, the field of reality TV is ripe for a more rigorous analysis.

In this paper, we apply advanced machine learning techniques to predict the most important features that determine the winner of a given season. In Section \ref{sec:data}, we describe the data used in this study. In Section \ref{sec:ML}, the three machine learning models used in this study are compared. And finally, in Section \ref{sec:conc}, we describe the implications of this paper and future work.


\section{Contestant Observations \& Data}\label{sec:data}

\begin{table*}[t]\centering
  \caption{Sample of Dataset}
  \begin{tabular}{cccccccccc}
    Name (Season) & Age & Hometown & Job Category & Race & 1-on-1 week & FIR? & Place \\ \hline
Michelle Kujawa (14) & 26 & West & Finance & White & \dots & No & 12 \\
Kirpa Sudick (23) & 26 & West & Dentist &  Asian & 5 & No & 5\\
Tia Booth (22) & 26 & South & Medical & White & 4 & No & 4\\
Sharleen Joynt (18) & 29 & Canada & Art & Asian & 3 & Yes & 6\\
Brooke Burchette (17) & 25 & East & Politics & Black & \dots & No & 17\\
Olivia Caridi (20) & 23 & South & News & White & \dots & Yes & 8
  \end{tabular}

 \begin{tablenotes}
      \small
      \item Table 1 is published in its entirety in the machine-readable format.
      A portion is shown here for guidance regarding its form and content. The full table can be found at 
      {\color{blue} \url{https://github.com/chesmore/bach-stats/}}.
    \end{tablenotes}
    \label{tab:data}
\end{table*}

We began by compiling our testing and training data sets. 
``Observations'' of episodes were conducted over a six-month period, from June to December 2021, and were collected on a 12-inch MacBook Pro using the streaming service \textit{Hulu}.  Seeing conditions were often hazy, due to the observers consuming multiple glasses of wine per night during data collection.

We compiled data from Seasons 11 to 25, for a total of 422 contestants. Season 11, Brad Womack's iconic first season in which he chooses no woman in the end, is often considered the beginning of the modern Bachelor franchise. Data for seasons prior to 11 were difficult to find, as they aired before the modern age of the internet. 
Therefore, analyzing data prior to Season 11 is beyond the scope of this paper.
Analysis of the \textit{The Bachelorette} is also left to the next paper in this series.

\begin{figure}
    \centering
    \includegraphics[width = .36\textwidth]{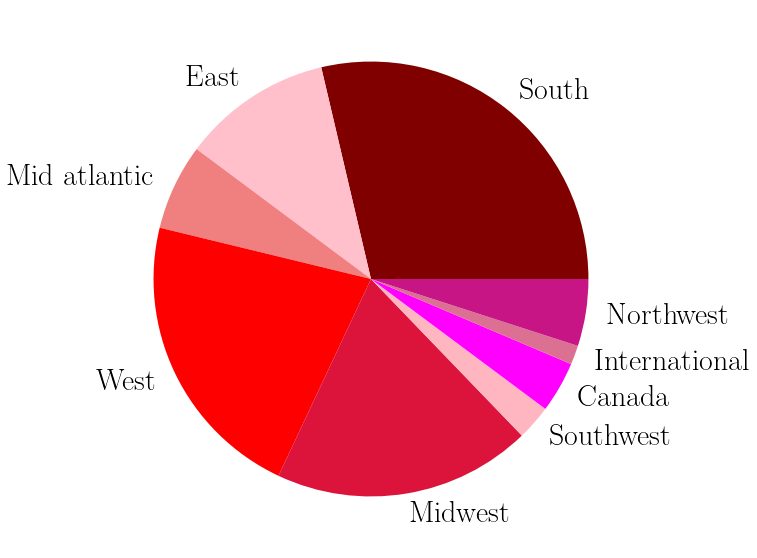}
    \includegraphics[width = .36\textwidth]{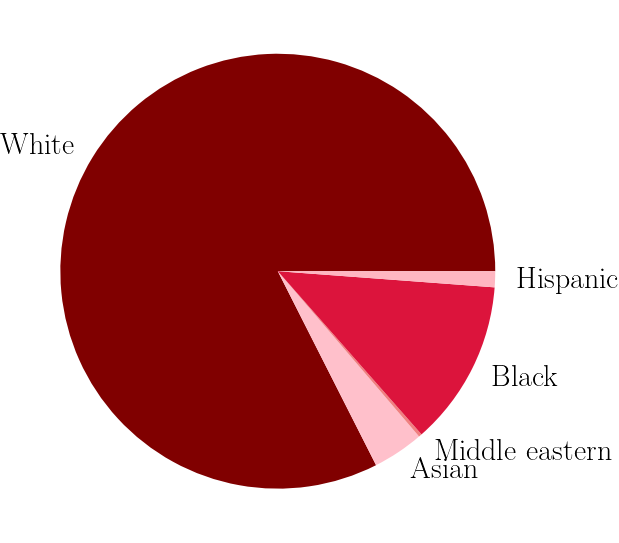}
    \caption{Top: Distribution of contestant Hometowns. The South is the most highly represented hometown region from \textit{The Bachelor}, making up about 29\% of contestants.  Bottom: Distribution of contestant race. White women overwhelmingly dominate the contestants from \textit{The Bachelor}, representing about 82\% of contestants.}
    \label{fig:ht}
\end{figure}

We trained our machine learning models on six contestant features: Age, Hometown location, Job, race, whether they got the First Impression Rose, and which week they got their One-on-One date. The label assigned to each contestant is then the place they received, where 1 corresponds to the contestant winning the season and higher values correspond to earlier eliminations. In Table \ref{tab:data}, we show a sample of our data. 

\begin{figure}
    \centering
    \includegraphics[width = .45\textwidth]{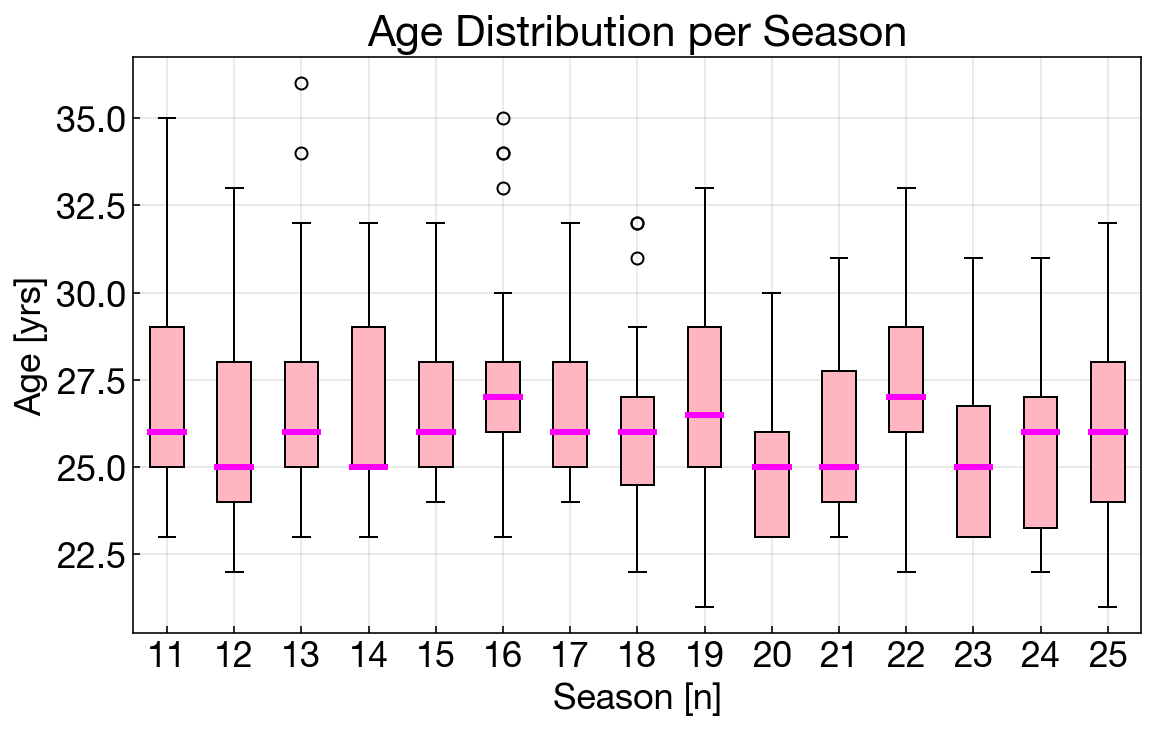}
    \caption{Box-and-whisker plot showing the ages of \textit{The Bachelor} contestants in the dataset.  There is no significant variation in mean age throughout the last 14 seasons, though we note there are fewer outliers in seasons 19 through the present than in previous seasons.}
    \label{fig:age_dist}
\end{figure}

Figure \ref{fig:ht} shows a distribution of the hometowns represented on this show.
Women from the South are the dominant population. 
Figure \ref{fig:ht} also shows the race distribution of contestants. Unsurprisingly, white women dominate the competition, a fact that has been long a source of controversy for the show.  
We also compiled data on contestant careers.
Notably, in 3\% of cases, contestants had eccentric job titles, likely for added attention. In these instances, we categorized their job category as `joke.' For example, Tiara Soleim (season 20) listed her occupation as `Chicken Enthusiast' (she was eliminated almost immediately). Lucy Aragon (season 18) listed her occupation as `Free spirit,' and placed 14th her season. Kelly Travis (season 18) listed her occupation as `dog lover,' and placed 9th.  
Figure \ref{fig:age_dist} shows the age distribution of the contestants in this analysis over the past 14 seasons of The Bachelor.  The distribution shows a consistent age-casting of approximately 26 years old.  

\begin{figure*}[t]
    \centering
    \includegraphics[width = .45\textwidth]{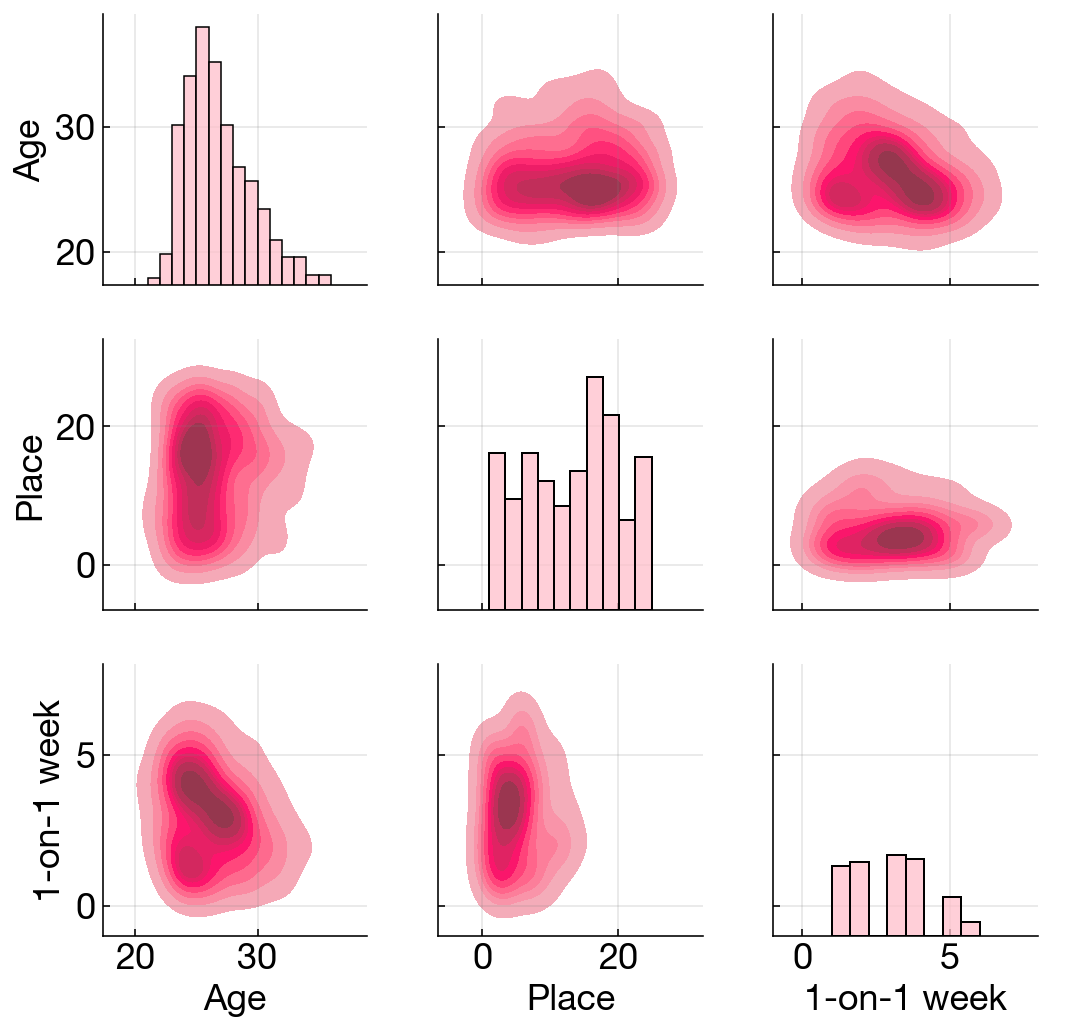}
    \includegraphics[width = .45\textwidth]{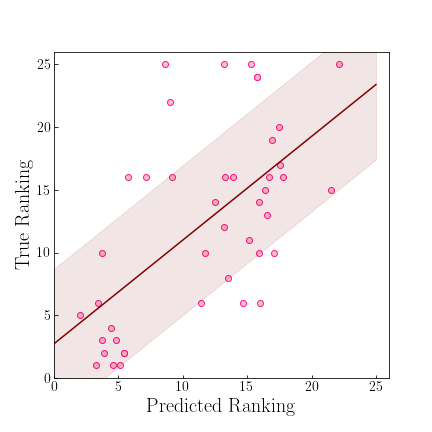}
    \caption{Left: Pair plot for age, final place in the season, and 1-on-1 week.  This plot demonstrates the high correlations between (1) 1-on-1 week vs. final place in the season; a late-week 1-on-1 date is correlated with higher success in the season, and (2) age vs. final place in the season; being in your mid-20s is often a predictor of success.  We also note that receiving a 1-on-1 date at all guarantees the contestant will finish the season as one of the top 15 contestants.  Right: Comparison of the True vs. Predicted ranking for our test dataset, using the Random Forest Classifier. The maroon line is the fit to points, with its 1-$\sigma$ uncertainty shaded in pink.}
    \label{fig:vars}
\end{figure*}

\section{Predicting Success with Machine Learning}\label{sec:ML}

In this section, we describe the three machine learning models applied to our data: random forest classification, neural nets, and linear regression. We comment on the relative success and utility of each model as applied to reality TV dating in Section \ref{subsec:compare}.

\subsection{Random Forest}
We used the random forest algorithm \cite{2001MachL..45....5B} from the package \texttt{Scikit-Learn} \cite{scikit} to classify a contestant's ranking. 
Random forest is a supervised learning algorithm that uses a forest of decision trees which each seek to find the best predictive splits to subset the data. Because there are multiple uncorrelated decision trees, averaging over many trees reduces the variance of the learning method.
In this work, we `label' contestants by the place they receive in the show.
We split our data into training/testing datasets, with 380 training points and 42 testing points, respectively. Data were one-hot encoded to allow the machine learning model to interpret categorical data better.

We trained on all the features discussed in Section \ref{sec:data}. Using the \texttt{sklearn.ensemble.RandomForestRegressor} module, we trained our classifier using a 20-tree model.
We found relative success given that our training dataset only included 380 datapoints. We define the accuracy rate as:

\begin{equation}
    \textrm{Accuracy} = 100 - \frac{\sum\limits^{N} (|\textrm{predicted} - \textrm{true}|/\textrm{true} \times 100)}{N},
\end{equation}

where N is the length of our test dataset, and predicted and true are the predicted ranking and true ranking of the contestants, respectively. Using this formula, we calculated a 30.5\% accuracy rate for our Random Forest Classifier model applied to our testing data. In Figure \ref{fig:vars}, we show a scatterplot comparing the true ranking vs. the predicted ranking for our test dataset.

To find the `ideal' contestant, we then fed our Random Forest model every possible combination of features (322,560 possible combinations). The simulated contestant with the highest ranking as predicted by our model is then defined as the `ideal' contestant. We found that the `ideal' contestant:

\begin{enumerate}
    \item 1-on-1 week: Received a 1-on-1 during Week 6.
    \item Race: Is White.
    \item FIR: Did \textit{Not} receive a First Impression Rose. 
    \item Job Category: Works as a Dancer.
    \item Hometown: Is from the Northwest.
    \item Age. Is 26 years old.
\end{enumerate}

We found the contestant that most closely matched this profile was Kaitlyn Bristowe who placed 3rd on season 18. Bristowe would go on to become a Bachelorette the next year.  



\subsection{Neural Network}
Using \texttt{pytorch} \cite{paszke2017automatic}, we created an artificial neural network (NN) to classify a contestant's ranking (Place) using data from Table \ref{tab:data}, modified in two ways.
Categorical data (Race, Hometown, Job Category, Joke Entrance) were transformed into one-hot encoded features, and both Hometown and Job Category were subdivided into broader categories instead of treating each unique Job or Hometown as a separate feature, to prevent over-fitting.
For example, we combined the Midwest, East, and Mid Atlantic Hometowns into the North East.
The only unnormalized numerical inputs we included were Age, Season, and 1-on-1 Week. 
We also removed 37 instances of contestants who finish in $16^\mathrm{th}$ Place, as they were over represented in our data set, causing our network to over-fit.

Our final NN architecture consisted of a single hidden layer, with a Leaky ReLU (Rectified Linear Unit) activation function and batch normalization in the hidden layer, and a Softmax activation function for the output layer.
We split our data set into $2/3$ training and $1/3$ for testing, and used the \textit{Adam} optimizer with a Cross Entropy loss function (standard for multi-class classification problems) to train the network.

\begin{figure}
    \centering
    \includegraphics[width = .5\textwidth]{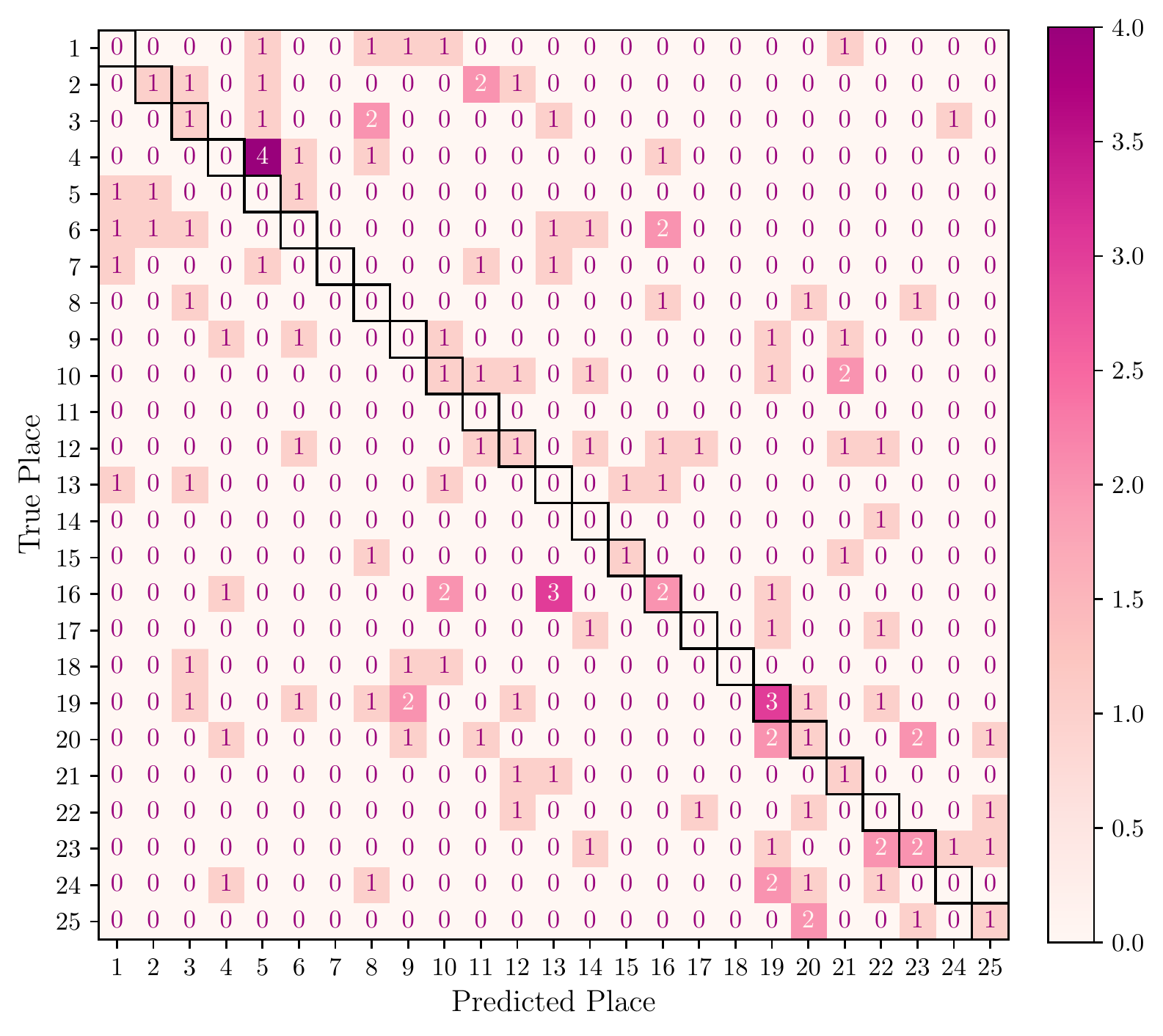}
    \caption{Confusion matrix of predictions from our neural network for the Place a given contestant will finish.
    We trained on $2/3$ and tested with $1/3$ of the data from Table \ref{tab:data} with some modifications to accommodate our classification neural network.
    Our network reached a balanced accuracy score of about $10\%$ and was able to correctly classify the Places of 15 contestants in the test set.
    However, as shown, our network could not correctly predict the $1^\mathrm{st}$ Place winner.
    }
    \label{fig:NN_conf_matrix}
\end{figure}

Figure \ref{fig:NN_conf_matrix} shows the confusion matrix from our NN predictions for the test set, which achieved a balanced accuracy score of $\sim10\%$.
However, we were not able to correctly predict the $1^\mathrm{st}$ Place winner, indicating that our neural network was not deep enough to parameterize the complex connections contestants forge over the course of the season. 
Although our network outperformed random classification, which achieved a balanced accuracy score of $4\%$, we expect that our performance was limited by the relatively small data set.

Training NNs with small data sets is challenging as the network tends to over-fit the data and has high-variance gradients, making optimization difficult \cite{liu2017deep}.
In Figure \ref{fig:NN_conf_matrix} we see the classes with the most correct predictions (Place 16 and 19) were also the two most represented classes within the data set.

In attempts to improve the performance of our network, we found that removing the Season from our data set reduced the balanced accuracy to $3\%$.
Therefore, it is important to consider how \textit{The Bachelor} has evolved as a franchise when extrapolating from training data in different seasons.

\subsection{Regression Machine Learning Models}

Finally, we modeled contestant performance using simple regression models from the \texttt{Scikit-Learn} Python package \cite{scikit}. We compared two regression models: a simple linear regression model and a Gaussian Process regression model, which assumes a subset of the data follows a Gaussian shape. Furthermore, because regression models generally do better with fewer features, the only two features that we trained on in this section were age and which week the contestant received a 1-on-1 date, which we found led to the best predictions of final ranking.

First, we trained the ordinary least-squares linear regression model. 
An ordinary least-squares linear regression fits a line through the training data data points, minimizing the sum of the squares between the fit and the training set.  This outputs an array of weights $w_i$ (Eq.~\ref{eq:weights}) which are used in the model to predict the outcome $\hat{e}$  (Eq.~\ref{eq:linreg}).  To train the model, we split the data into 2/3 training and 1/3 for testing.

\begin{equation}
    \text{min}_\omega ||X\omega - e ||^2
    \label{eq:weights}
\end{equation}
\begin{equation}
    \hat{e}(\omega,x) = \omega_0 + \omega_1 x_1 ... \omega_n x_n
    \label{eq:linreg}
\end{equation}

Second, we implemented a Gaussian Process Regression.
We found the Gaussian Process regression yielded a higher R value than a Linear regression, predicting the success of a contestant with an R value of 0.35.  

We concluded from the GPR model that the later a contestant receives a 1-on-1 date, the higher they place in the overall season.  
For example, if a contestant receives a 1-on-1 in the final few episodes ($>7$), they are predicted to finish in fourth place, or better.

\subsection{Comparing Machine Learning Models as Applied to Reality TV Dating Shows}\label{subsec:compare}

One of the goals of this paper was to find which machine learning algorithm was the most successful at predicting success for \textit{The Bachelor} and reality TV dating shows in general. We found the neural network to have the highest balanced accuracy score, although future implementation may be difficult because selecting the hidden layers can be complex. The regression algorithm was simple to employ, but may be less ideal for predicting complex phenomena like reality TV romance, as regression models are generally more successful with fewer features. In conclusion, we generally recommend neural nets for future reality TV dating show strategical studies; random forest classifiers are slightly less effective but simpler to implement overall.

Across all three models, we found that receiving a 1-on-1 is a good indicator that a contestant will place at least in the top 10. 
Some features were difficult to fully explore due to their limited representation in the data set. For example, the percentage of People of Color (POC) contestants was too small to make accurate statements about correlations between race and success on \textit{The Bachelor}.  We hope that The Bachelor will mitigate this problem, both for statistical analysis purposes and for general enjoyment of the show.

\section{Conclusion \& Discussion}\label{sec:conc}

\subsection{Recommendations for \textit{The Bachelor} Producers}

\begin{figure}
    \centering
    \includegraphics[width = .5\textwidth]{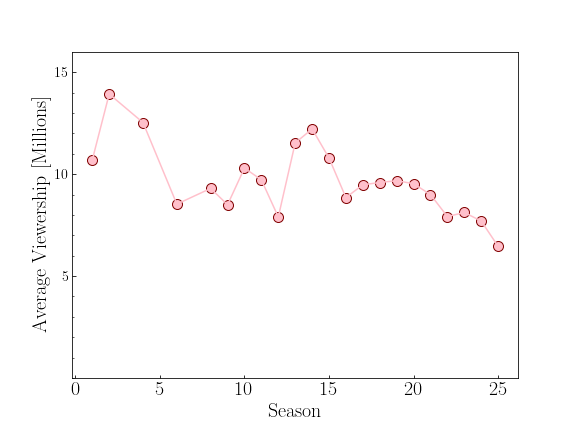}
    \caption{Average viewership in millions of \textit{The Bachelor} over the show's 25 first seasons. The average rating has consistently declined. We lay out a set of recommendations to producers in Section \ref{sec:conc} for how to improve the show's format.}
    \label{fig:rate}
\end{figure}

The authors of this paper have been long-time `mega-fans' of \textit{The Bachelor}. Figure \ref{fig:rate} shows the steady decline in ratings of \textit{The Bachelor}, particularly in the last five years. 
The authors believe this is caused by two main factors: (1) Contestants increasingly are more motivated by the prospect of becoming `instagram influencers' than by finding love. This often leads to highly manufactured drama as contestants scramble for more screen time, even to the detriment of their romantic connection with the lead. (2) The production crew increasingly meddle in the storylines (e.g., `Champagne Gate of season 24). This can cause viewers to feel like the show is scripted and predictable. 

As self-proclaimed mega-fans, we have developed a set of recommendations for the \textit{The Bachelor} production crew to increase ratings again. We recommend the following:

\begin{enumerate}
    \item Switching up the format of each season so that contestants are less likely to come in with a strategy. Season 19 of \textit{The Bachelorette} is doing just that, with two bachelorettes instead of one. 
    \item Only casting non-influencers. This could be done; for example, by only casting contestants without instagrams, or putting in contestant contracts they are not allowed to sell products for at least two years after filming. This will incentivize contestants to focus more on the love than becoming famous.
    \item Randomly selecting contestants for \textit{Bachelor in Paradise}. Often contestants will be overly dramatic so they can secure a spot on \textit{Bachelor in Paradise}. If contestants are selected randomly, this will remove the incentive for contestants to participate in fake and manufactured drama.
\end{enumerate}

\subsection{Concluding Remarks}

In this work, we present three machine learning models that predict the success of a contestant on \textit{The Bachelor}. We found that the best-performing model was our neural network, which we recommend for future studies regarding reality TV show predictions.

Higher accuracy rates may have been achieved by larger testing data sets. We plan to continue this work by adding new data as each subsequent \textit{The Bachelor} season is released.  Therefore, our models will only continue to improve in predicting a contestant's outcome.
That being said, our models are hampered by more than just data availability; True Love cannot be predicted.

The software and data used in this analysis are public on GitHub and are available as a downloadable Python package, called \verb|bach-stats|~\cite{bachstats}.  
The module contains Jupyter Notebooks which recreate figures from this work.

\section*{Acknowledgment}
The authors would like to thank ABC and \textit{The Bachelor} for providing them with joy and laughter during their PhDs. AJL would like to thank Ashley Villar and Vanessa Böhm for useful discussions.

AJL, JM, KAR, and MS thank the LSSTC Data Science Fellowship Program (DSFP) for giving them the machine learning skills to produce such an impactful and field-changing work as this one. This project was developed in part at the 2022 LSSTC DSFP Hack Day session.  GEC is supported by the National Science Foundation Graduate Student Research Fellowship (Award \#DGE1746045), though this work was done during GEC's free time.

\bibliographystyle{IEEEtran.bst}

\bibliography{sample}

\end{document}